# Rolling control and dynamics model of two section articulated-wing ornithopter

Guangfa Su[1], Yu Cai[1*], Jiannan Zhao[1*]

[1]Guangxi Key Laboratory of Intelligent Control and Maintenance of Power Equipment, School of Electrical Engineering, Guangxi University, Nanning 530004, China

**Abstract:** This paper invented a new rolling control mechanism of two section articulated-wing ornithopter, which is analogues to aileron control in plane, however, similar control mechanism leads to opposite result, indicating the ornithopter supposed to go left now go right instead. This research gives a qualitative dynamics model which explains this new phenomenon. Because of wing folding, the differential rotation of outer-section wing (analogues to aileron in plane, left aileron up and right aileron down make left turn) around pitch axis becomes common mode rotation around yaw axis, leading its rotating torque changing from left-handed rotation (using left-handed as example, right-handed is the same) around roll axis to a common mode force pointing to front-right (northeast, NE) direction from first player's view of the ornithopter. Because most of the flapping movement is in the upper hemisphere from ornithopter's view, the NE force is above on the center of mass of the orthopter, generating a right-handed moment around roll axis. Therefore, the ornithopter supposed to go left now goes right. This phenomenon is a unique and only observed in two section articulated-wing ornithopter by far. Many field tests conducted by authors confirm it is highly repetitive.

## 1. Introduction

Articulated-wing ornithopter was invented by Festo and made its successful flight in 2011 [1]. Its commercial name is "Smartbird" and becomes popular around the world in very short time because its flying kinematics is very similar to a real bird. Although its kinematics had been discussed in many research papers [2], its dynamics model had not been explicitly explained and rarely discussed. The wings of articulated-wing ornithopter consist of inner wing and outer wing. In the paper [3,4], it is mentioned that the inner wing is used to provide lift and flapping outer wing is used to generate thrust. Steering is through tail, similar to rudder and horizontal tail in plane, although not exactly the same. Two servos mounted in the tip of each outer wing to achieve active torsion, mainly used to enhance thrust efficiency [3]. Flapping wing dynamics model of non-articulated wing, or called single section wing, is similar to a propeller changing its rotating direction periodically, switching between counterclockwise and clockwise back and forth, with its pitch changing between positive and negative in the same rhythm. This mode can generate an air flow toward the same direction, out of which thrust comes consequently. Pitching movement of most of the single section wing is through difference of rigidity between the front end of the wing membrane and the rear end of wing membrane. During flapping movement, the front end of the wing membrane is with high rigidity and the rear end is with low rigidity, so that the rear end would have more deformation under air pressure and with larger rotation angle around the wing spar than front end, pitching movement of wing achieved thereafter. This is called passive torsion. Similar to helicopter, which can change the pitch angle of the propeller to achieve higher thrust. Servos mounted on the outer wing tips in Smartbird can adjust pitch angle of the outer wing to achieve active torsion, generating more thrust compared to legacy passive torsion. Optimized active torsion of outer wing can enhance thrust efficiency, but required servos knows the phase and in which stage that the inner wing and outer wing in flapping cycle, and also required to be with high torque and being able to twist wing periodically and synchronized with inner wing and outer wing flapping movement. Servos added in the outer wings also increase the inertia of the wing, making its structure more vulnerable compared to passive torsion.

The outer wing of Smartbird is made by highly flexible membrane [1,4], inner wing is covered by thin foam skin and skeleton is made by carbon fiber. The cross section of the inner wing is an airfoil [3,4]. It is relatively more rigid than outer wing. Outer wing can make passive torsion even without servos assistance like many non-articulated wing ornithopter. However, the legacy research and papers didn't mention whether servos added in the outer wing tips would have any impact on ornithopter's attitude in flight.

**2. Motivation**

Compared to the dynamics model of non-articulated wing, articulated-wing and plane, it is found that the plane can be analogues to a rigid body, skeleton of non-articulated wing model analogues to single pendulum and skeleton of articulated wing model analogues to double-pendulum (Fig. 1). The difference between ornithopter dynamics and legacy rigid body dynamics is that the flapping wing aerial vehicle dynamics has to take the interaction between wing membrane and air into account. The dynamics of articulated wing becomes mixture of multi-rigid body dynamics and aerodynamics, though the question is not easy because the most of ornithopter's wing is flexible and cannot be modeled as a rigid cardboard. Even if cardboard model is valid, due to non-steady flow generated by flapping movement, the exact force asserting on the cardboard is difficult to calculate quantitatively.

Legacy non-articulated flapping wing kinematics is superposition of rooted-flapping movement around shoulder axis (parallel to roll axis) and pitching movement around wing spar. However, the double pendulum model of the articulated flapping wing movement indicates that although the inner wing movement similar to non-articulated wing, the outer wing kinematics is much more complicated and how to model its interaction with the air is still open.

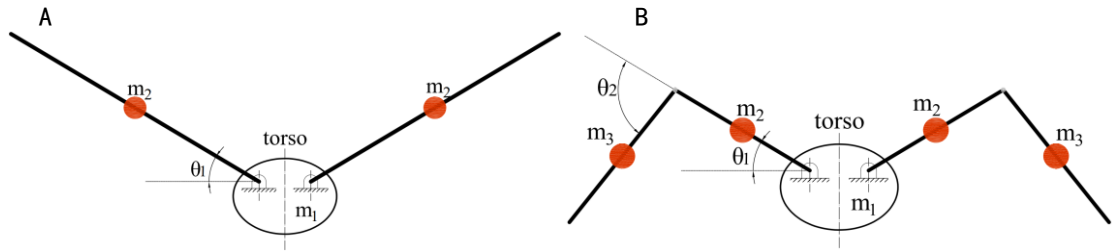

Fig. 1 Skeleton model of non-articulated wing and articulated wing. (a) The single pendulum model of single section wing. (b) The double pendulum model of the articulated two section wing.

This paper gives a qualitative explanation of articulated-wing rolling control mechanism based on experimental phenomena and double pendulum cardboard model. The motivation of this research comes from the counter-intuitive result of the articulated flapping wing rolling control mechanism. Similar rolling-control mechanism in articulated-wing ornithopter, non-articulated wing ornithopter and plane leads to opposite steering result in flight.

### 3. Flapping articulated wing kinematics

The flapping wing kinematics of the ornithopter developed by the authors (Fig. 2) is similar to Smartbird [7]. The skeletons of the outer wing and inner wing are connected by two four bar linkage. The movements of the outer wings and inner wings are coupled and not independent, so that they are not 2 degree of freedom fully actuated system.

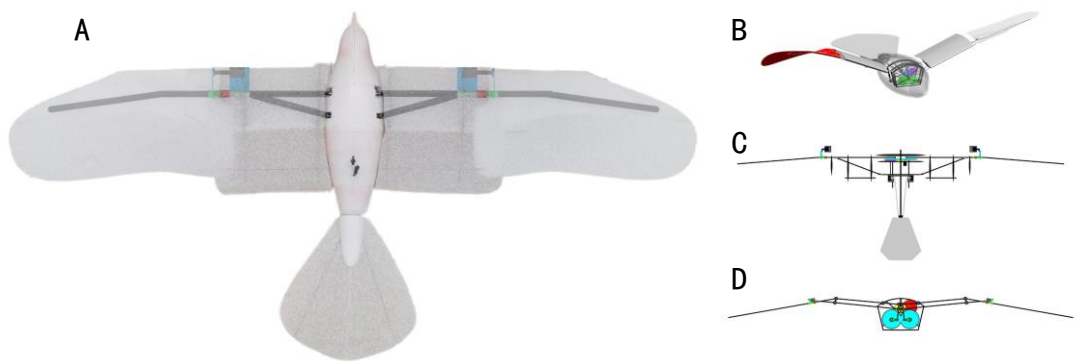

Fig. 2 Flutterplane construction diagramTwisted wing structure(a) Position of the twist mechanism on the flutterplane(b) 3D modelling model of a flapping wing aircraft(c) Top view of the internal frame of the flapping wing(d) Front view of the internal frame of the flapping wing.

The prototype of articulated wing ornithopter was developed by authors' team on which all experiments are conducted. The original version does not have twisted wing structure in the outer wing. Steering of the ornithopter is via yaw movement of the tail, similar to rudder in plane. Tail is connected to torso via a universal joint and it can move vertically and horizontally to generate yaw moment or pitch moment, though some yaw moment can couple to roll moment when the tail pitch angle is large, though the coupling effect is not significant. Steering can be achieved by horizontal movement of the tail. Steering by tail is highly effective during cruise flight, for example, when pitch angle of the ornithopter is small. If the pitch angle of the ornithopter is too large, steering becomes difficult. The wings have to be extremely symmetric due to lack of rolling control mechanism. Lack of symmetry between wings might lead to low manoeuvrability or slipping drift in steering. The highly symmetric wing fabrication is indispensable for successful flight of the ornithopter.

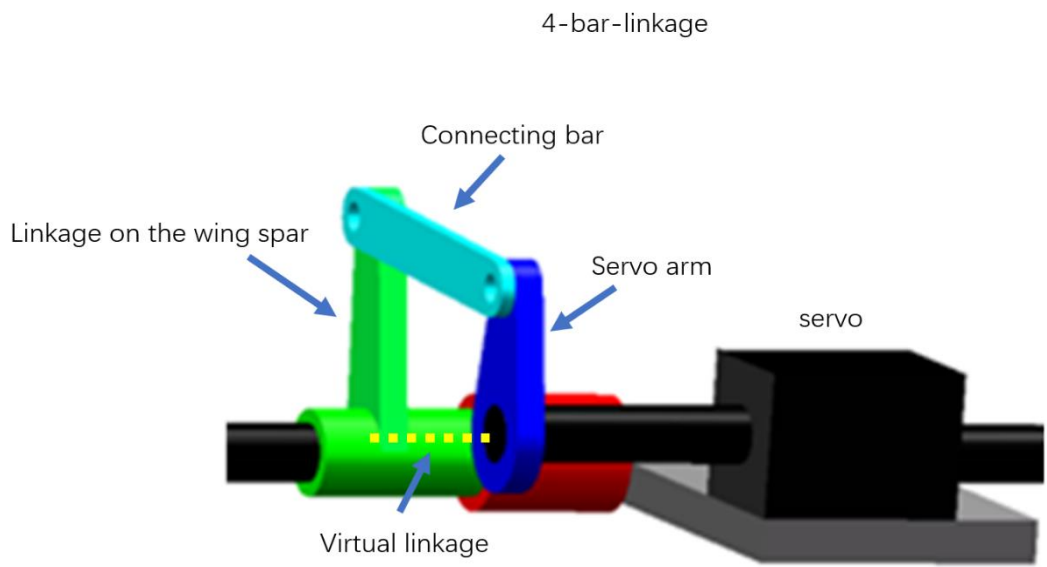

Fig. 3 Twisted wing structure, a four bar linkage. Servo arm (blue), linkage on the wing spar (green), connecting bar (light blue), virtual linkage between wing spar and root of servo arm.

Authors developed a twisted wing mechanism (Fig. 3) where servo is mounted on the root of the outer wing to enhance its steering capability. The servo arm connects a bar mounted on a wing spar via a linkage. The servo arm, linkage on the wing spar, connecting bar and a virtual linkage between servo arm root and wing spar forms a four bar linkage. Wing spar can rotate around its axis by servo actuation. There is a bend in the wing spar of outer wing (Fig. 2), which can twist wing membrane when the rotation occurs in the wing spar root. The wing rib (normally a airfoil shape) is glued in the outer wing membrane and doesn't move when the wing spar rotate. Servos in the left wing and right wing rotate in differential mode, leading the wing membrane twists in differential mode. Video in supplementary shows how it works. Different from Smartbird where servos are mounted on the tips of outer wing, servos in this paper placed close to the joint between inner wing and outer wing. Servos are fastened on the root chassis of outer wing. The functionality of the servos is to change the neutral flapping position of outer wing rather than doing active torsion. The twisted wing mechanism is quite similar to rolling control mechanism in plane where ailerons position change in differential mode. The difference between articulated ornithopter and plane is that the aileron of the articulated-wing ornithopter becomes the whole outer wing rather than small part of the wing in plane. Flapping outer wings generate most of thrust in ornithopter while plane's thrust comes from propeller or the turbine engines.

Field test found out that under the twisted wing mechanism, the rolling moment appears and the difficulty of steering due to lack of wing symmetry alleviated. If neutral position of tail selected properly, the ornithopter can steer by twisting wing differentially without assistance of the tail. Field test results shocked the authors that, this twisted wing mechanism leads to a steering result contrary to what was predicted based on the legacy plane or non-articulated wing ornithopter model (Fig. 4).

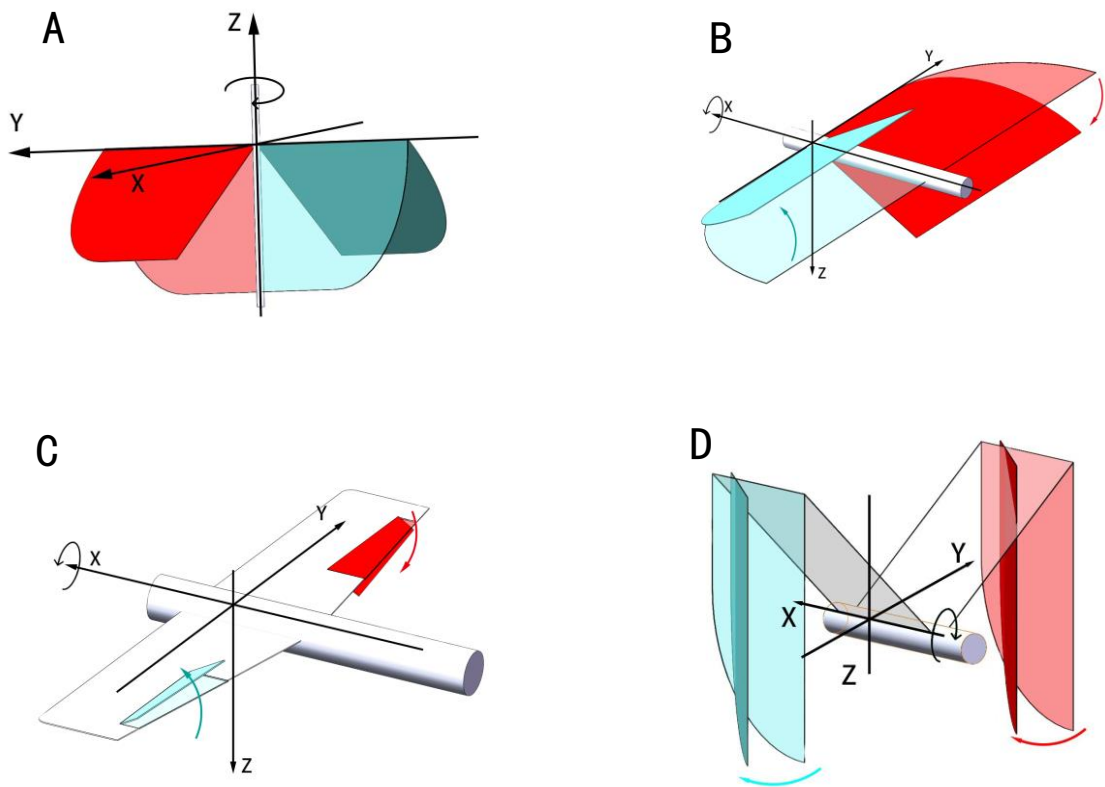

Fig. 4 Differential twisting wing mechanism in different aerial vehicles (using left aileron up and right aileron down as example)  (a)  Left-handed rolling in hovering non-articulated wing ornithopter (b)Left-handed rolling of non-articulated wing ornithopter during cruise flight (c) Left-handed rolling of plane (d) Right-handed rolling in articulated wing ornithopter.

## 4. Result and explanation

Field test data in Fig. 5 shows the ornithopter makes rolling movement. Its twisted wing mechanism is contrary to aileron twisting in plane, indicating if left turn required, the left aileron of the ornithopter (left outer wing) should go down (twisted down) and right outer wing goes up (Fig. 6). Up is the direction toward upper hemisphere or zenith direction. The phenomenon is highly repetitive and observed many times in field tests. The data was collected and recorded by Radiolink mini pix [8] flight controller with Ardupilot firmware [9].

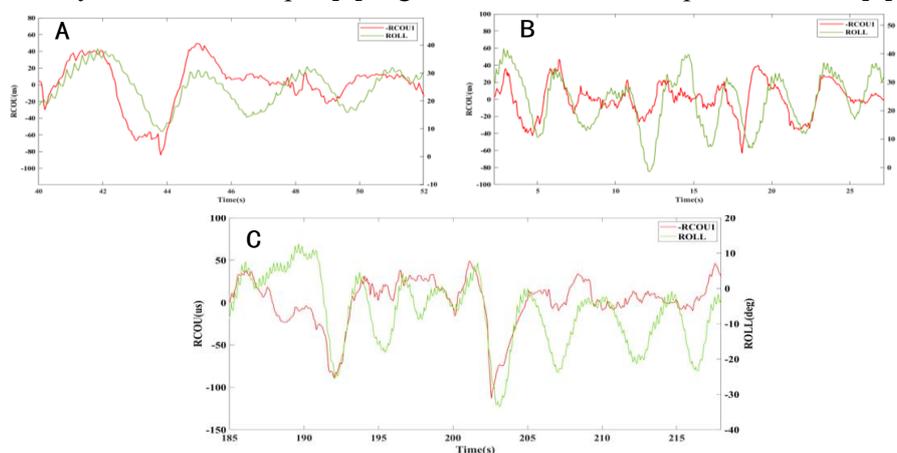

Fig. 5 Relationship between the roll angle measured by the attitude sensor and the control

signal

Authors gives a qualitative dynamics model to explain this phenomenon (Fig. 7) The articulated wing flapping model in Fig. 7 shows a model in extreme condition where the outer wings folded to vertical position so that the wings are in a capital M shape. Although this folding is exaggerated and not realistic, it can give an intuitive understanding why the left-handed torque around roll axis becomes right-handed.

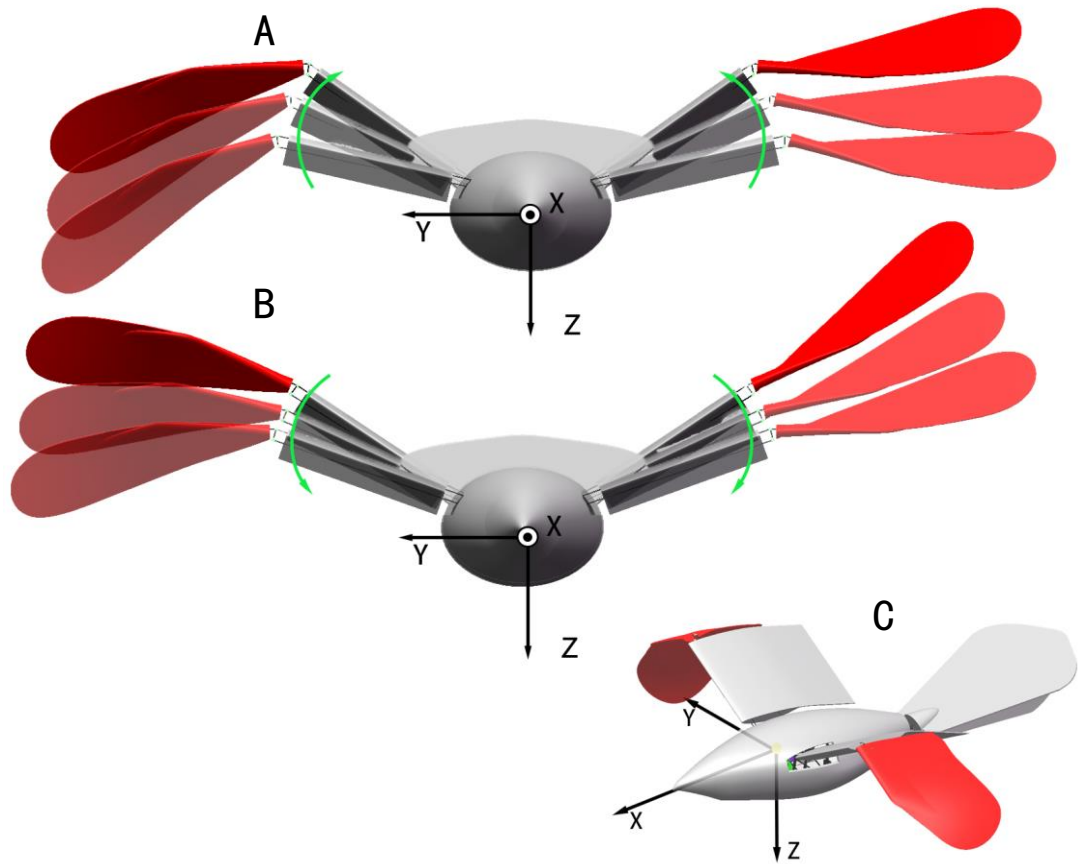

Fig. 6 Contrary to plane, left outer wing twists up and right outer wing twists down generate right turn movement. (A) Up stroke (B) Downstroke (C) side view

Because of folding of outer wing, the original twisting rotation is around pitch axis (wing spar) now becomes twisting around vertical axis (parallel to yaw axis) (Fig. 7). Using left-hand rolling in the legacy design as example, the aileron in left wing goes up and right aileron goes down in a plane lead to left-hand rolling. Similarly, the twisting of wing around wing spar in the same manner leads to left-hand rolling in non-articulated single wing ornithopter, either in hovering flight or in cruise flight (Fig. 4). However, in articulated-wing ornithopter, the twisting wing after folding is around vertical axis (parallel to yaw axis), both outer wings now point to the front-right (northeast) direction from ornithopter first player's view (Fig. 7). The thrust vectors generated by flapping wing change their directions because neutral flapping position changed. The assertion points of the thrust generated by flapping outer wing is above center of mass (CoM) of the ornithopter and their direction is toward northeast. That is because during one cycle of flapping wing of both inner wings and outer wings, most of phases of flapping movement is in the upper hemisphere from ornithopter's point of view. The center of flapping thrust or lift is above the center of mass of the ornithopter as well. Just like most of the capital

M in Fig. 7 is in the upper hemisphere. Therefore, generated moment become right-hand moment around roll axis. That's why the similar twisted wing mechanism leading to opposite steering result.

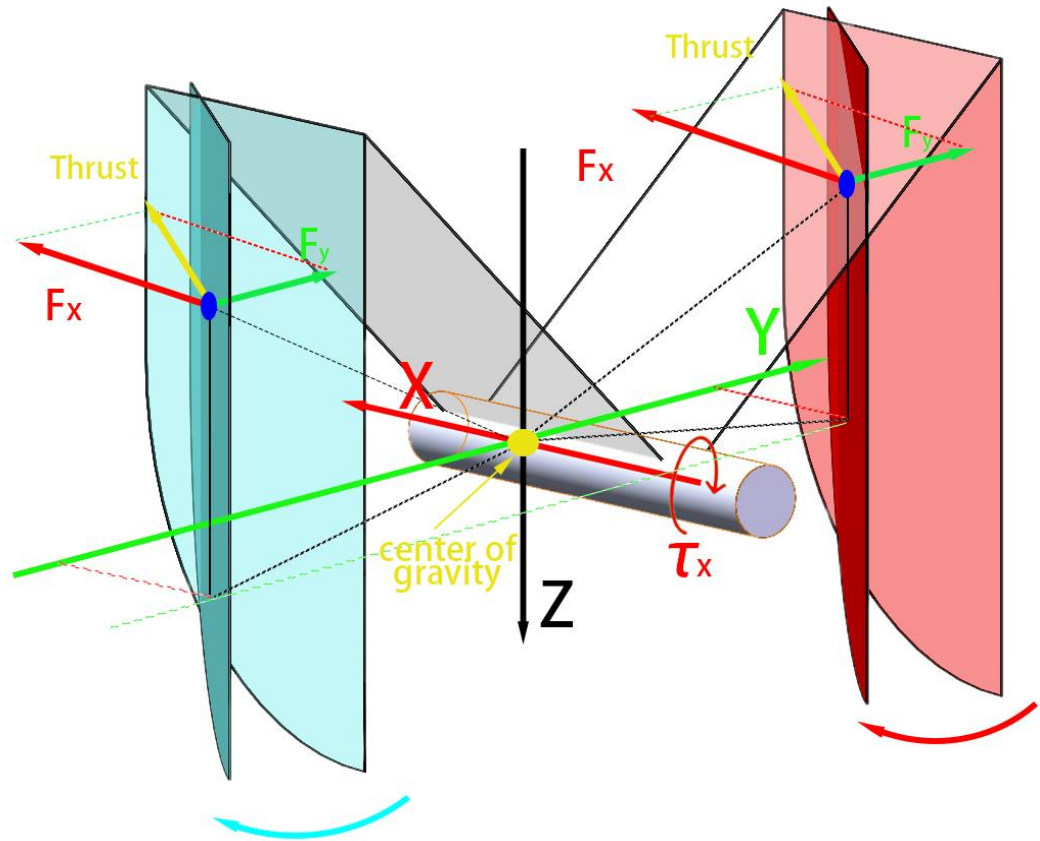

Fig. 7 Differential twisted wings becomes common mode twisted wings due to wing folding

Above models explains why the articulated-wing ornithopter has this counter-intuitive phenomenon. Folding kinematics of the outer wings changed thrust vectors from original differential mode to common mode. Both outer wings thrust vectors now point to northeast (common mode) due to left-wing going up and right-wing going down (differential mode).

In reality, the kinematics of outer wing is much more complicated than the model description and the outer wing would not bend to vertical position in any case. Fig. 7 uses an extreme case to explain how folding mechanism changes twisted wing mode and associated rolling moment. Therefore, even the outer wings do not bend to vertical position, it does not change the physics of its right-handed moment generation mechanism because the thrust vectors has components pointing to right and their assertion points are above CoM, thus right-handed rolling is inevitable.

## 5. Conclusion and Future work

This paper only gives a qualitative explanation to this abnormal phenomenon. Its explicit quantitative dynamics model is still required to do more investigation in the future. Because the wing membrane is flexible rather than rigid and current model is based on rigid cardboard model. Modeling damping effect of the air on the cardboard is also challenging. How to combine the multi-rigid body dynamics with non-steady aerodynamic effects still requires more attentions in the future work.